\documentclass[conference]{IEEEtran}
\usepackage{times}
\usepackage[T1]{fontenc}
\let\labelindent\relax
\usepackage{enumitem}
\usepackage{amsmath}
\usepackage{amssymb}
\usepackage{graphicx}

\usepackage[numbers]{natbib}
\usepackage{multicol}
\usepackage[bookmarks=true]{hyperref}

\usepackage{ulem}
\usepackage{color}

\newcommand{\edit}[2]{#2}

\begin{document}

\title{A Single Motor Nano Aerial Vehicle with Novel Peer-to-Peer Communication and Sensing Mechanism}

%
\author{\authorblockN{Jingxian Wang\authorrefmark{4},
Andrew G. Curtis\authorrefmark{4},
Mark Yim\authorrefmark{2},
and Michael Rubenstein\authorrefmark{4}}
\authorblockA{\authorrefmark{4}
Center for Robotics and Biosystems,  Northwestern University}
\authorblockA{\authorrefmark{2}GRASP Lab, University of Pennsylvania}
}

\maketitle

\begin{abstract}
Communication and position sensing are among the most important capabilities for swarm robots to interact with their peers and perform tasks collaboratively. 
However, the hardware required to facilitate communication and position sensing is often too complicated, expensive, and bulky to be carried on swarm robots.
Here we present Maneuverable Piccolissimo 3 (MP3), a minimalist, single motor drone capable of executing inter-robot communication via infrared light and triangulation-based sensing of relative bearing, distance, and elevation using message arrival time.
Thanks to its novel design, MP3 can communicate with peers and localize itself using simple components, keeping its size and mass small and making it inherently safe for human interaction.
We present the hardware and software design of MP3 and demonstrate its capability to localize itself, fly stably, and maneuver in the environment using peer-to-peer communication and sensing.

\end{abstract}

\IEEEpeerreviewmaketitle

\section{Introduction}
\label{sec:introduction}

Uncrewed aerial vehicles (UAVs) have been extensively studied in recent years for their wide applications in photography \cite{DJIDrones}, exploration \cite{slam}, transportation \cite{lift}, and drone shows \cite{traj}. UAVs that collaborate with each other can usually improve performance in these tasks, enabling the swarm to carry heavier objects \cite{multiload1,multiload2,multiload3}, perform more complicated actions \cite{actions1,actions2}, explore faster \cite{explore2}, or form more intricate shapes \cite{periodic,droneshows}.

In order to collaborate, UAVs in the swarm need inter-robot communication and position sensing capabilities \cite{abdelkader2021aerial}. In most systems, these capabilities are provided externally, by a WiFi router and an optical tracking system \cite{actions2,periodic,crazy}, Lighthouse systems \cite{drew}, or RTK GPS \cite{rtkgps}. Thus, these systems are usually limited to the particular environment in which they operate. There have been attempts to solve this problem by sensing using on-board sensors \edit{, via either a}{like} depth cameras \cite{swarm1}, Lidars \cite{swarmlio}, or less accurate ultra wide band (UWB) modules \cite{uwb1}. However, these modules, especially depth cameras and Lidar, are often expensive, heavy and bulky, not to mention that they also require powerful processors to process the data generated \cite{swarm1,swarmlio}. The complexity and cost associated with manufacturing and maintaining such systems can limit the scalability of a swarm \cite{pcbot}.

Additionally, a wide variety of tasks, including search and rescue, transportation, or drone shows, can involve direct or indirect interaction with humans. But highly capable multi-motor drones tend to be heavy and unsafe for human interaction without safety equipment \cite{unsafe}, a risk that is amplified when interacting with a large number of UAVs. While there are impressive works on active avoidance of obstacles which partially addressed this issue \cite{avoid1,swarm1}, it would be preferred if the drone could be light enough to be inherently safe \cite{harmless} to interact with.

\begin{figure}[tbp]
    \centering
    \includegraphics[width=\linewidth]{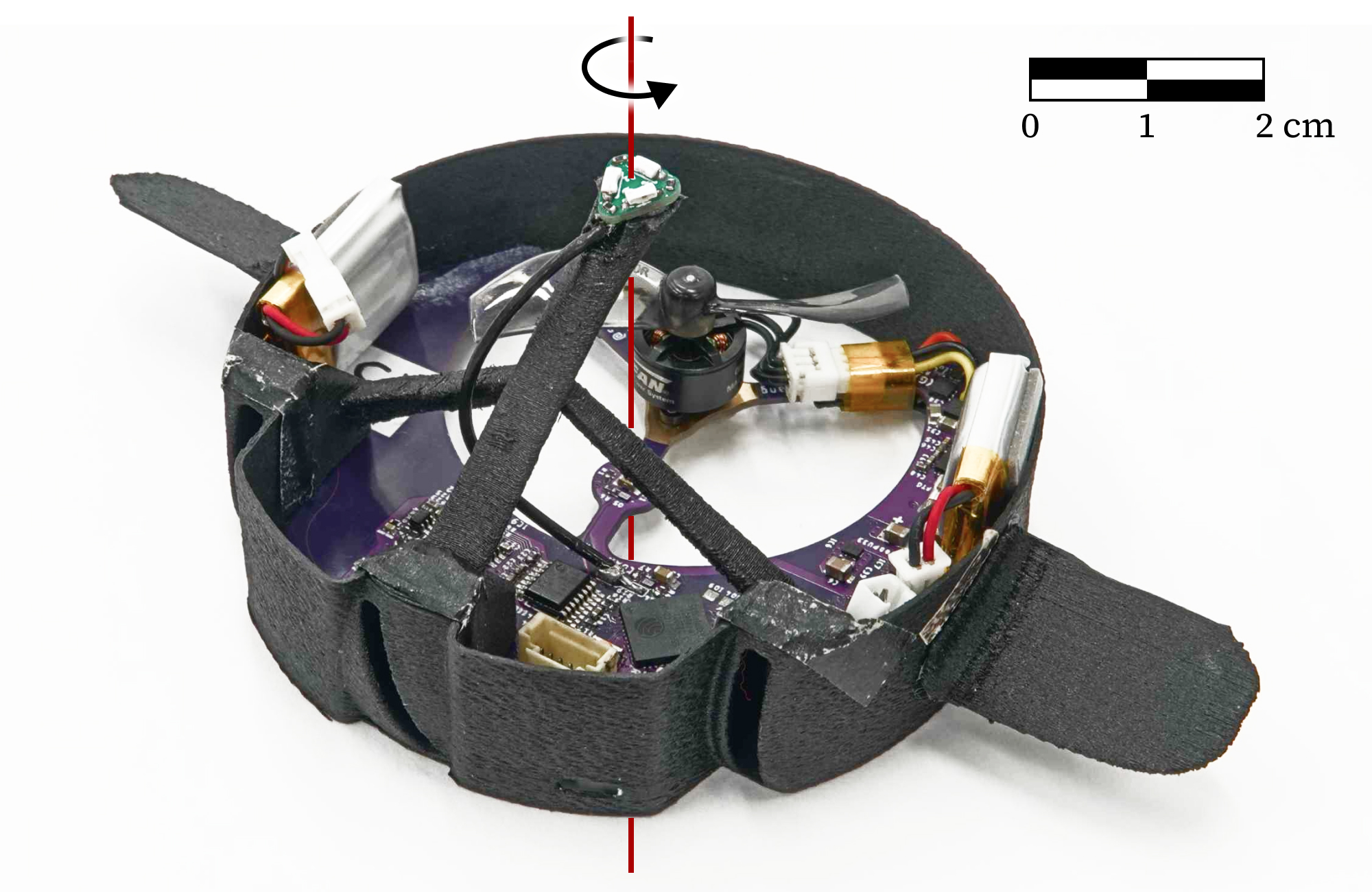}
    \caption{A photo of MP3. MP3's size is $\varnothing 112\times38$mm and weighs 20.0g. As MP3 flies, the whole robot rotates around the geometric axis indicated in the figure.}
    \label{fig:drone}
\end{figure}

There is often a trade-off between simple, lightweight drones and drones with advanced capabilities. However, recent efforts in single actuator flyers \cite{mono1,mono2,msam,pico,drew} have overcome some of the flight and control limitations of simple drones while maintaining reduced complexity and weight. An interesting feature of these single-motor drones is that their bodies spin when flying. As the body rotates, sensors placed on the body can scan through the environment. This feature could be exploited to achieve omnidirectional communication and sensing \cite{sense1,coaxial} while maintaining a simple, lightweight form factor.

In this paper, we present a single motor nano \cite{nanodef,crazy} aerial vehicle, MP3, with a novel, minimalist peer-to-peer communication and sensing system. MP3 is similar to its predecessors: its only motor is offset from its center of mass to provide maneuvering ability similar to the original Maneuverable Piccolissimo (MP) \cite{pico} and its controller is similar to MP2 \cite{drew}. However, MP3 is equipped with a novel communication and sensing system as well as a completely new chassis, electrical, and software design. MP3 is capable of communicating with neighboring beacons or other MP3s using infrared light (IR), and sensing its relative bearing, distance, and elevation to neighbors using a single-drone triangulation method. Using this information, MP3 can subsequently determine its global Cartesian position with millimeter accuracy. Using the global position information, MP3 can control its position in 3D space with a single actuator. The simplicity of MP3 allows it to be 20.0g in weight and 112mm in diameter. The small mass and the partially enclosed propeller make MP3 inherently safe for human interaction even with collisions \cite{harmless}. We present the hardware and software design of MP3, with emphasis on the novel communication and sensing system, and validate its capability with experiments.

\begin{figure}[htbp]
    \centering
    \includegraphics[width=\linewidth]{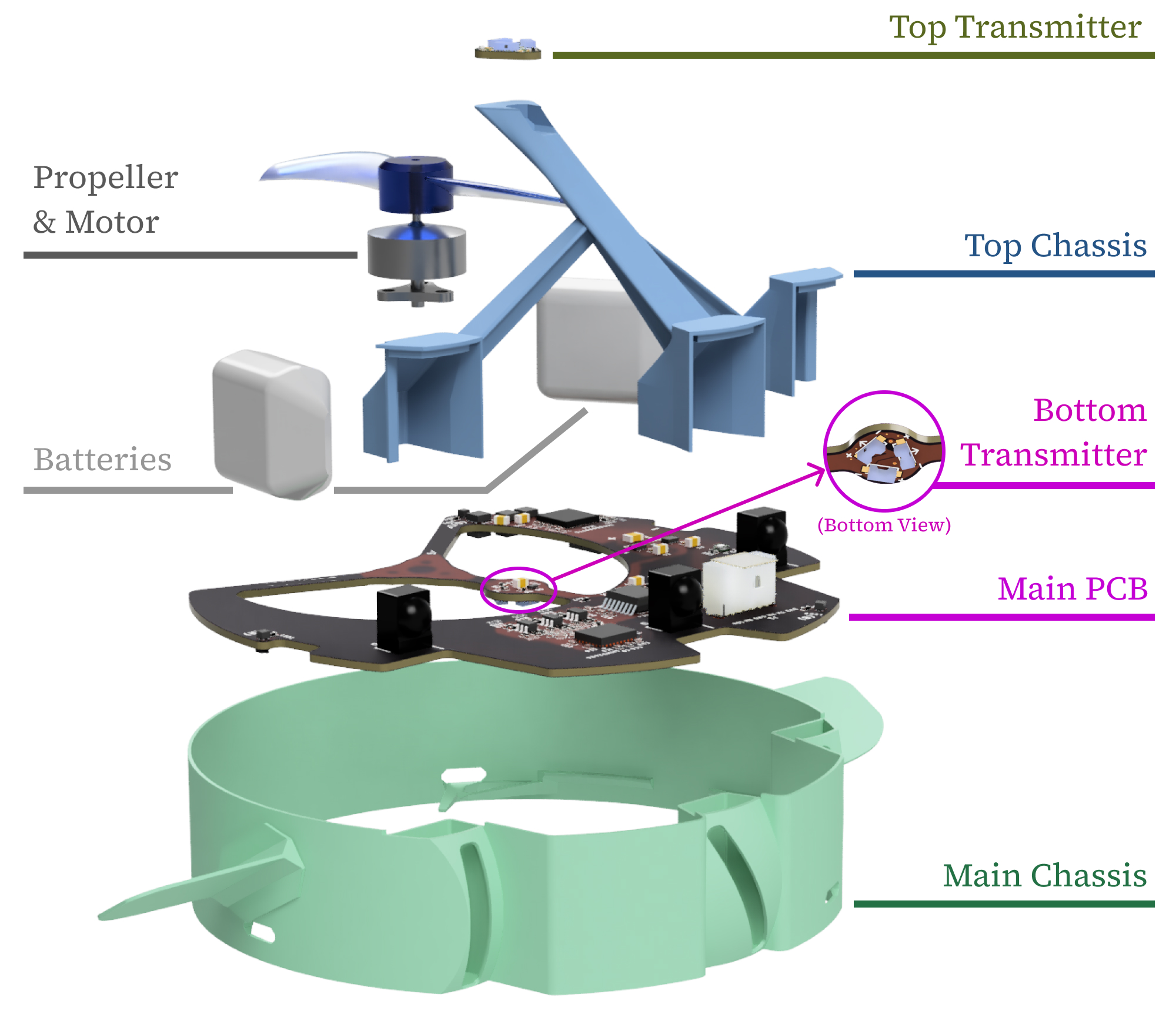}
    \caption{An exploded view of MP3. MP3 has 7 major parts: 3D printed main and top chassis, a main PCB, a top transmitter, a motor-propeller assembly, and two batteries. The bottom transmitter is placed at the center of the bottom side of the main PCB.}
    \label{fig:exploded}
\end{figure}

\section{Robot Design}
\label{sec:robotdesign}

\subsection{Overall Concept}
\label{sec:concept}


MP3, as shown in Fig. \ref{fig:drone} and \ref{fig:exploded}, is a minimalist single-rotor drone with on-board communication and position sensing capabilities using a novel optical communication and sensing system. Unlike ordinary drones, MP3 has only one motor offset from its center of mass, two rigid bodies (the rotor fixed to a propeller and the stator fixed to the main chassis), and no inertial measurement unit (IMU). As the motor spins, the body of MP3 spins like a Frisbee in the opposite direction as shown in Fig. \ref{fig:drone}, and the gyroscopic effect of the spinning body keeps the MP3 stable. By pulsing the thrust at different times, we can tilt the MP3 and control its 3D position in flight \cite{drew}.

\begin{figure}[htbp]
    \centering
    \includegraphics[width=\linewidth]{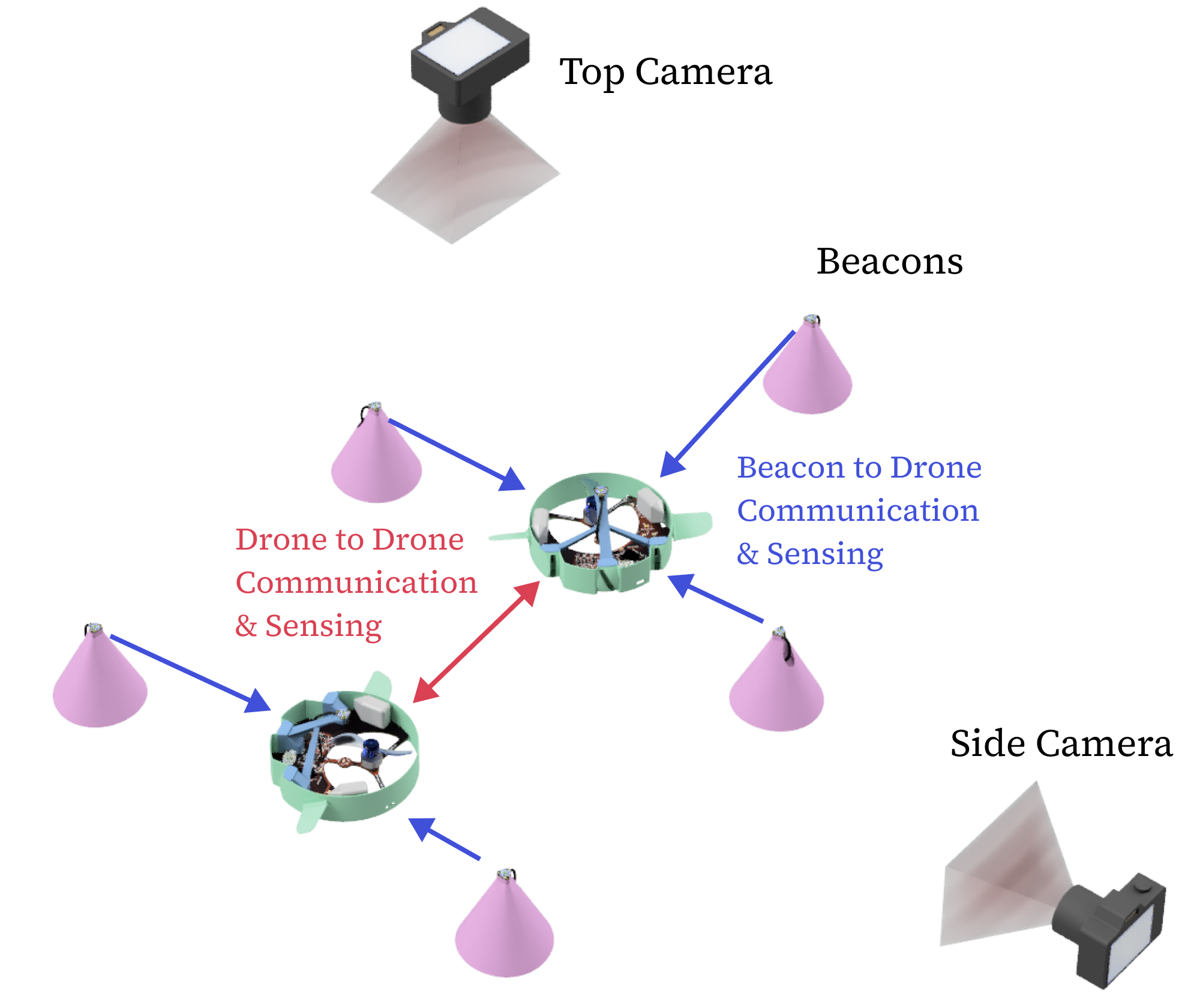}
    \caption{The setup of the test environment. Several beacons and MP3s are in the environment, communicating and providing others with localization. Two cameras, one on the top and one from the side, capture the trajectory of the drone, providing us with the drones' reference trajectory for analysis only. Blue arrows indicate uni-direction transmission from the beacons to MP3s, and red arrows indicate bi-directional communication between MP3s. Figure is not drawn to scale.}
    \label{fig:setup}
\end{figure}

\begin{figure}[htbp]
    \centering
    \includegraphics[width=\linewidth]{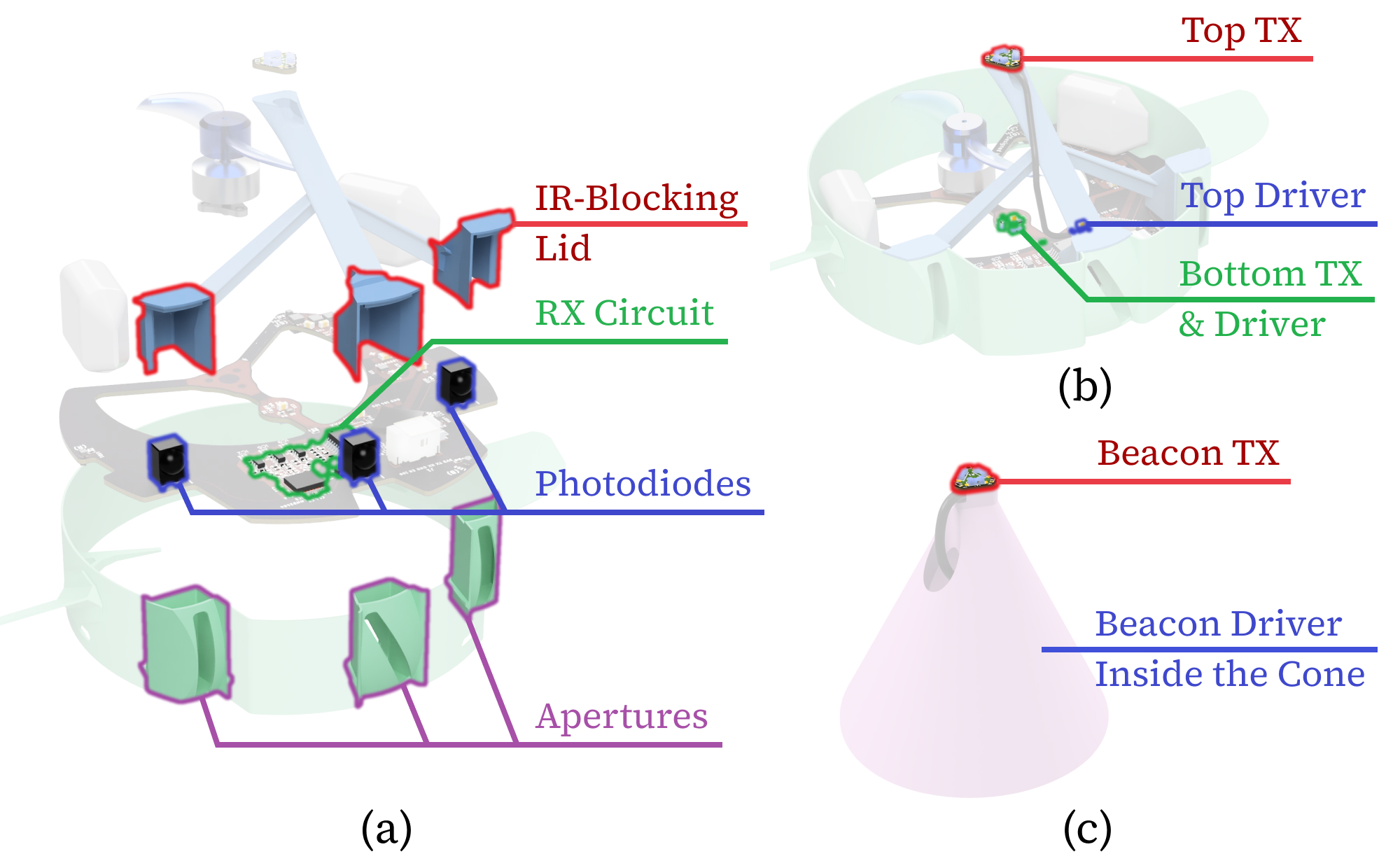}
    \caption{MP3's communication and sensing system. (a) shows the receiver system (RX) on MP3, (b) shows the top and bottom transmitter system (TX) on MP3, and (c) shows the transmitter system on the Beacon. Note that the transmitter system on the MP3 and on the Beacon are the same.}
    \label{fig:comloc}
\end{figure}

MP3s operate in an environment with multiple MP3s and beacons, as shown in Fig. \ref{fig:setup}. In the environment, MP3s can communicate with and localize from other MP3s or beacons to execute tasks like position holding or waypoint following. It is worth noting that beacons share the same transmitter hardware and code as regular MP3s as shown in Fig. \ref{fig:comloc}, so beacons are essentially immobile MP3s that are constantly transmitting their position. \edit{}{By using beacons that have identical electronics to the drones, we can replace beacons with actual drones. In the future, we hope to replace all beacons with drones, enabling a swarm-based positioning without any external fixtures and achieving "infrastructure-free" operation as proposed by \citet{landandloc}. }We will use ``MP3s" to refer to both flying MP3s and beacons when talking about communication, sensing and localization in future discussions. \edit{The fact that beacons are immobile MP3s means we can replace beacons in the environment with flying MP3s, eliminating the need for having any infrastructure in the environment for the swarm to operate, and achieving "infrastructure-free" operation as proposed by Pourjabar et al. [31]}{}.

While the non-stationary body reference frame of the robot may seem to make things difficult (the MP3 body spins at \textasciitilde25Hz), we can actually take advantage of this property to achieve communication and sensing in all directions. As shown in Fig. \ref{fig:comloc}(a), there are three directional infrared (IR) photodiodes fixed to MP3's body acting as receivers. As MP3 spins, the receivers' field of view (FOV) scans through its environment, enabling it to communicate with MP3s all around it. Furthermore, the three receivers pick up signals from other transmitters at different times, and we can use this timing information to achieve bearing, distance, and elevation sensing. 

\subsection{The Communication and Sensing System}
\label{sec:commsystem}
A MP3 can communicate with other MP3s using packets of encoded IR pulses, and it can determine its relative position to nearby MP3s by measuring when its directional receivers start to receive packets from them. Compared to measuring relative distance using received signal strength, our timing-based approach does not require the transmitter and receivers to be uniform in intensity and sensitivity and is less likely to be influenced by scattered signals or changing environmental lighting conditions, making the system more reliable and robust.

Based on this principle, we designed the communication and sensing system hardware as shown in Fig. \ref{fig:comloc}. Omindirectional transmitters on MP3s blink their infrared (IR) LED to broadcast messages in the form of encoded pulses, and MP3s can receive messages and localize themselves using the three photodiodes on them.


\subsubsection{Theory of Operation}
\label{sec:theory}


\begin{figure}[htbp]
    \centering
    \includegraphics[width=\linewidth]{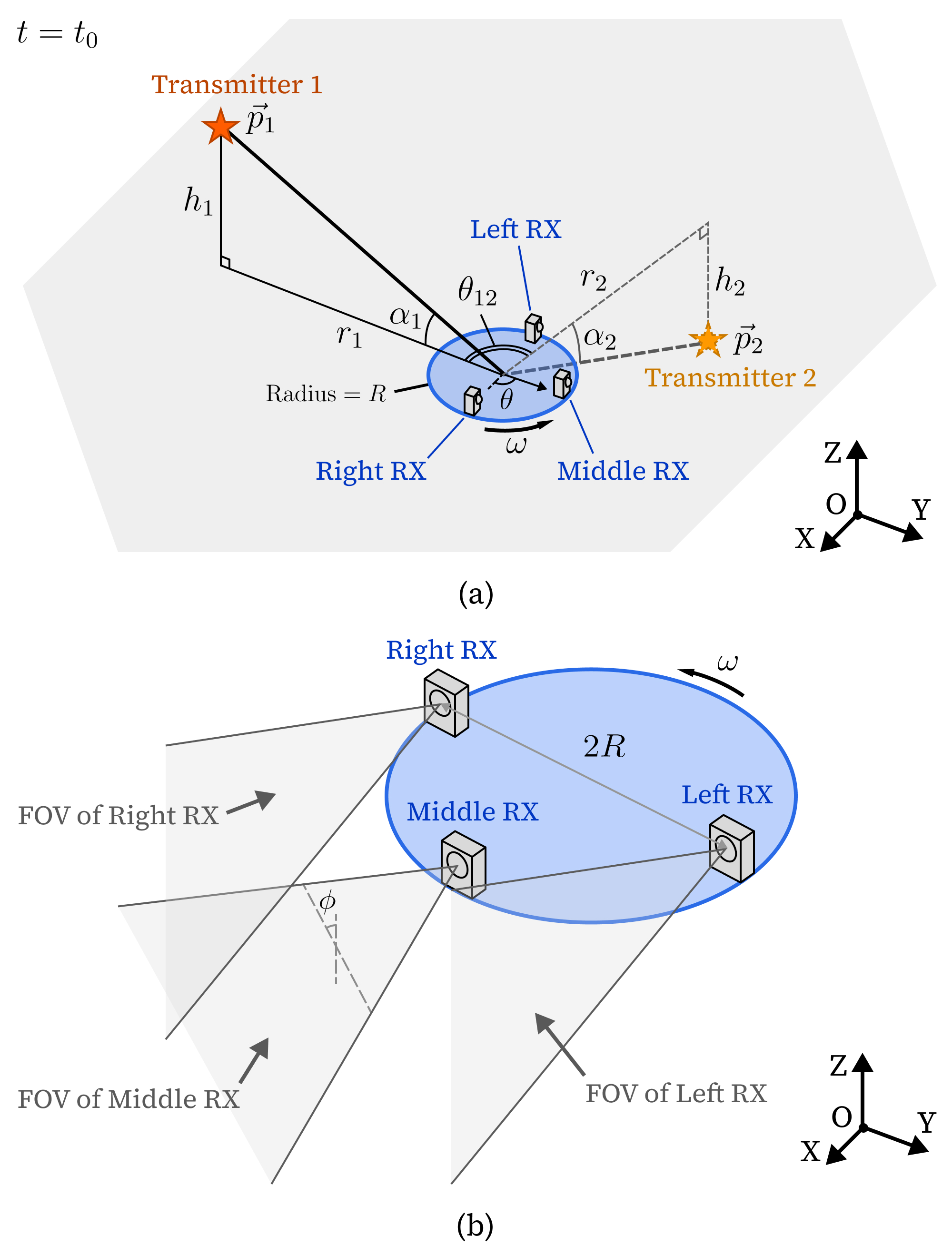}
    \caption{A simplified model of MP3. (a) shows the environment MP3 operates in and (b) shows the field of view of all three receivers.}
    \label{fig:model}
\end{figure}

To illustrate how MP3s achieve relative position sensing and absolute localization, we consider a simplified theoretical model as shown in Fig. \ref{fig:model}, where two point-like transmitters are constantly emitting light omnidirectionally and broadcasting their global position $\vec{p}_1,\vec{p}_2$ via IR communication. A simplified MP3 is trying to determine its relative bearing, distance, and elevation to the transmitters as well as its global position. The MP3 is modeled as a solid body rotating at a constant angular velocity $\omega \hat{z}$, and is equipped with three light receivers with planer fields of view (FOV), i.e. a receiver will only pickup light signal when the transmitter is in its FOV. Two light receivers (right and left) are oriented orthogonal to the MP3 horizontal plane, and one light receiver (middle) is oriented at an angle ($\phi$) from orthogonal. As the MP3 rotates, its left, middle, and right receiver will pickup signals from transmitter 1 at time $t_{\text{1L}},t_{\text{1M}},t_{\text{1R}}$ respectively, so we can determine the distance and elevation angle between the MP3 and transmitter 1 as:

\begin{eqnarray}
\label{eqn:rh1}
    r_1&=&\frac{R}{\sin{(\omega(t_{\text{1R}}-t_{\text{1L}})/2)}}\\
\label{eqn:rh2}
    \tan{\alpha_1}&=&\sin{(\omega(2\,t_{\text{1M}}-t_{\text{1R}}-t_{\text{1L}})/2)}\cot{\phi}
\end{eqnarray}

\noindent We can obtain $r_2, \alpha_2$ from transmitter 2 similarly. Furthermore, at time $t_1=(t_{\text{1L}}+t_{\text{1R}})/2$ and $t_2=(t_{2L}+t_{2R})/2$, the MP3 is facing transmitter 1 and 2 respectively, so we can express the angle between transmitter 1 and transmitter 2 as

\begin{equation}
    \theta_{12}=\omega(t_1-t_2)
\end{equation}

\noindent With this information, we could theoretically find the global position $\vec{p}$ of the MP3 by incorporating the calculated relative distance and elevation angles with the communicated absolute positions of the transmitters ($\vec{p}_1,\vec{p}_2$).

\edit{}{Furthermore, if we have a small uncertainty $\sigma_t$ for all timing measurements, we can also obtain uncertainty of $r_1$, $\alpha_1$, and $\theta_{12}$ as follows:}

\edit{}{
\begin{eqnarray}
\label{eqn:sh1}
    \sigma_{r_1}&\approx&\frac{r_1^2\,\omega\sigma_t}{\sqrt{2}R}\\
\label{eqn:sh2}
    \sigma_{\alpha_1}&\approx&\cot{\phi}\cos^2\alpha_1\,\omega\sigma_t\\
\label{eqn:sh3}
    \sigma_{\theta_{12}}&=&\sqrt{2}\,\omega\sigma_t
\end{eqnarray}
}

\noindent \edit{}{While $\sigma_{\alpha_1}$ and $\sigma_{\theta_{12}}$ are on the same order of magnitude as $\omega \sigma_t$ and are not very sensitive to relative position between robots, $\sigma_{r_1}$ is proportional to $r_1^2$, meaning that the localization is more accurate when drones are closer to each other. This feature is helpful for collision avoidance in dense swarms.}

Unfortunately, non-ideal factors make the implementation of this theory more complicated. These factors include non-ideal sensor optics, limited time resolution constraints by the communications protocols, etc.



\subsubsection{Optical Design}
\label{sec:optical}


When designing the optical system, we want it to be as close as possible to the ideal theoretical model of omnidirectional point-source transmitters and planar FOV receivers while keeping the design as simple as possible.

\begin{figure}[htbp]
    \centering
    \includegraphics[width=\linewidth]{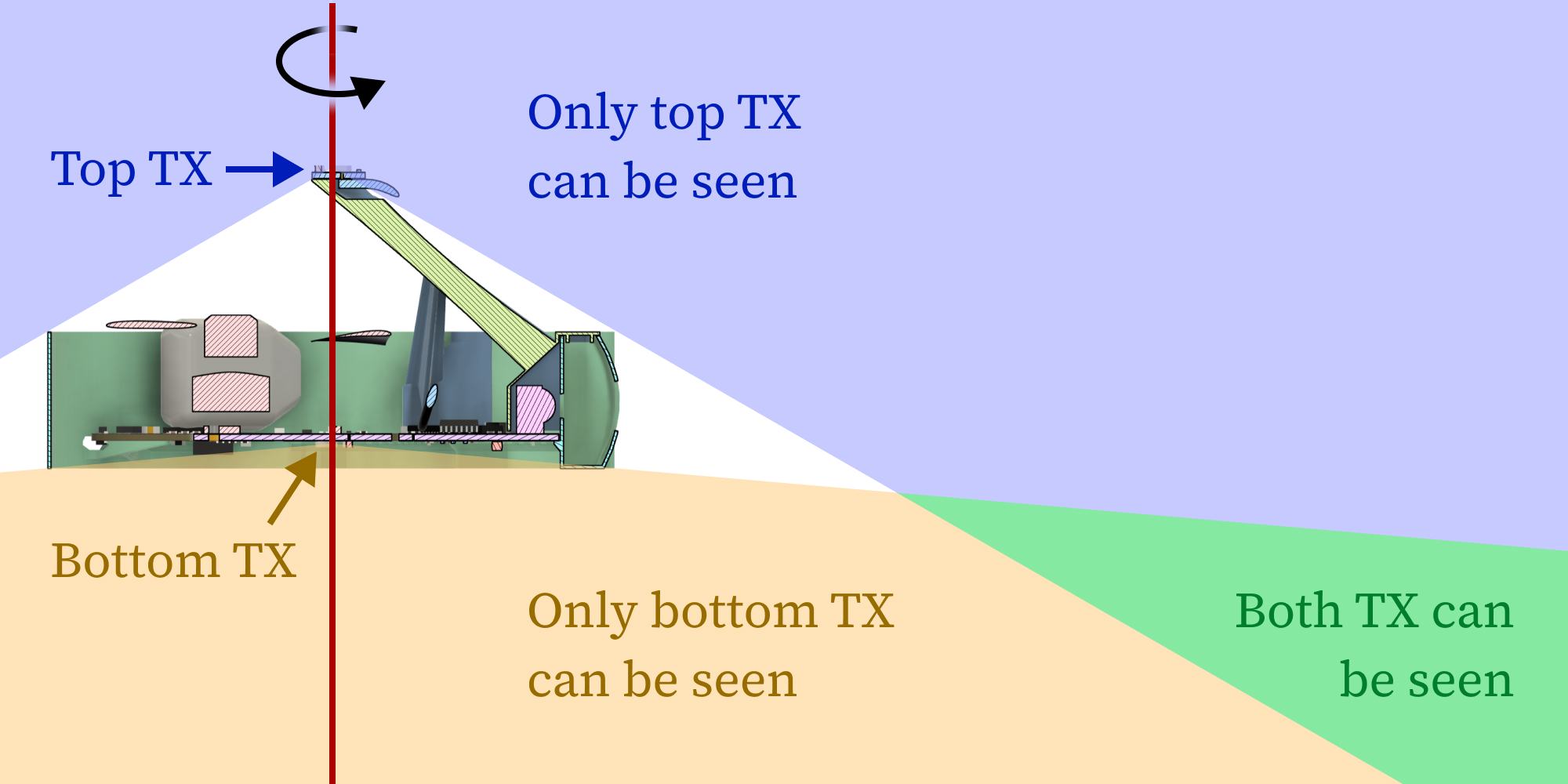}
    \caption{Transmitters' coverage region's cross section. Blue and orange regions are where the receiver on another MP3 can receive the signal from the top and the bottom transmitter respectively, and the green region is where both transmitters can be seen.}
    \label{fig:TX}
\end{figure}

For the transmitter, we used three closely placed side view LEDs, oriented 120 degrees apart, to approximate a omnidirectional point light source. Moreover, MP3s have a light-blocking, cylindrical-shaped chassis, so if we only place one transmitter on top of the MP3, other MP3s will not be able to receive the light when below it as the light will be blocked by the chassis. The condition is similar when we only place one transmitter at the bottom. To address this issue, we placed two transmitters on each MP3, one above the chassis and one at the bottom, increasing the region where the transmitter could be seen, as shown in Fig. \ref{fig:TX}. 

\begin{figure}[htbp]
    \centering
    \includegraphics[width=\linewidth]{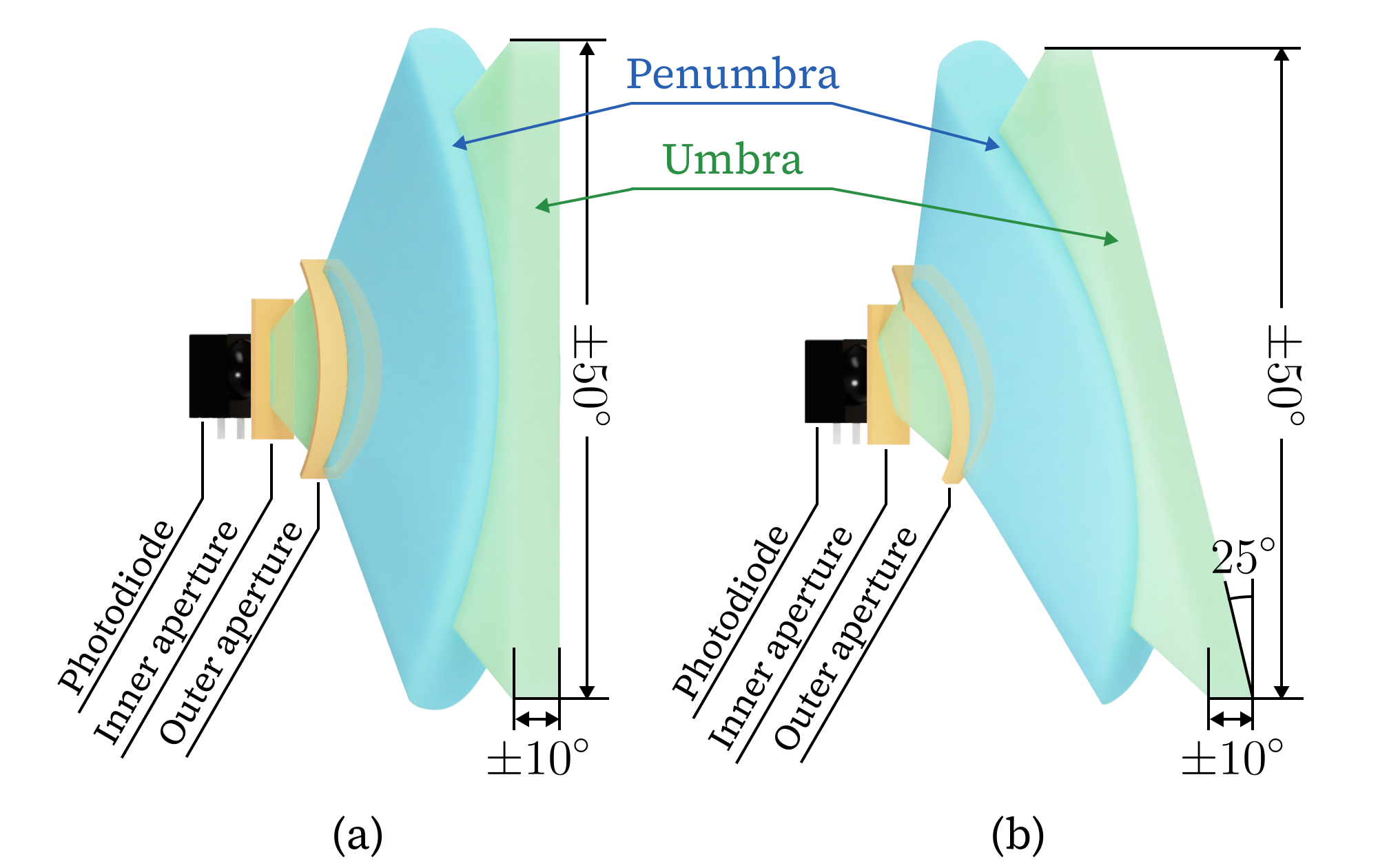}
    \caption{The receiver's optical structure. (a) shows the structure of the left and right receivers, and (b) shows the structure of the middle receiver. There are two apertures in front of each photodiode to restrict the FOV to $\pm 10^\circ$ horizontal and $\pm 50^\circ$ vertical. The middle receiver's aperture is tilted by $25^\circ$.}
    \label{fig:RX}
\end{figure}

For the receiver, we used three high sensitivity IR photodiodes to sense the IR signals, and each of them is placed behind two narrow linear apertures to create an approximate plane-shaped FOV. In reality we have finite aperture and sensor sizes, so the FOV is not a perfect plane; it has some thickness. So, instead of using the time when the receivers pick up the signal from the transmitters to compute the distance and elevation as mentioned in section \ref{sec:theory}, we now use the time when receivers \emph{start} to pick up the signal instead. The configuration of the apertures and the corresponding FOV are shown in Fig. \ref{fig:RX}.

Both the inner and outer apertures are important for creating a planar-like FOV. For the inner aperture, we want to increase its size to allow enough light to pass through. However, a finite-size inner aperture combined with a finite-size sensor creates a penumbra, where the light strength is reduced and varies depending on the direction. This introduces uncertainty about whether the signal will be received or not, which affects the accuracy of position sensing. Therefore, we also want to keep the inner aperture small to reduce this uncertainty. On the other hand, the outer aperture determines the FOV of the sensor, which influences how many transmitters could be seen at the same time and how long a transmitter will be in sight in a single rotation. These parameters are more important to communication and will be further explored in section \ref{sec:protocol}. Eventually, we set the inner aperture size to $1.25\times5$mm, horizontal FOV to $\pm10^\circ$, vertical FOV to $\pm 50^\circ$, and the tilting angle of the aperture on the middle receiver to $25^\circ$.

\subsubsection{Electrical Design}
\label{sec:electrical}

MP3 can only receive messages from others when its receivers are facing a transmitting MP3.
MP3s typically spin at 25Hz and each of their receivers have a $\pm 10^\circ$ horizontal FOV, meaning that, in each revolution, a MP3 only has approximately 2-5ms to communicate with another MP3\edit{}{, and in the worst case scenario this gives a communication latency of 40ms}. To facilitate communication with multiple MP3s using time division multiple access (TDMA), each MP3 could take no more than 500\textasciitilde 1000us to transmit all necessary information to a receiver. If we want to communicate tens of bytes over this period, we need a system capable of communicating at around 1Mbps. This means that we could not use commercially available IR transceivers but had to implement our own.

Heavily constrained by space, weight, power, and limited available hardware solutions for optical communication, we decided to use a relatively simple communication method: transmitters can only transmit information by turning on and off LEDs, and the receivers will pass the analog received signal through a comparator and convert it back to a binary signal before being processed by the on-board microcontroller. With these constraints in mind, our transmitter and receiver circuit are designed to facilitate high frequency (-3dB analog bandwidth of 10.5M) modulation and reliable reception of IR signals in multi-point to multi-point communication scheme over $5-50$cm free space.

Transmitters are 940nm IR LEDs driven by FETs, offering high peak power and fast response \cite{osram}. Each LED consumes 700mA when turned on, and has a rise and fall time of 8ns and 4ns, respectively.

\begin{figure}[htbp]
    \centering
    \includegraphics[width=\linewidth]{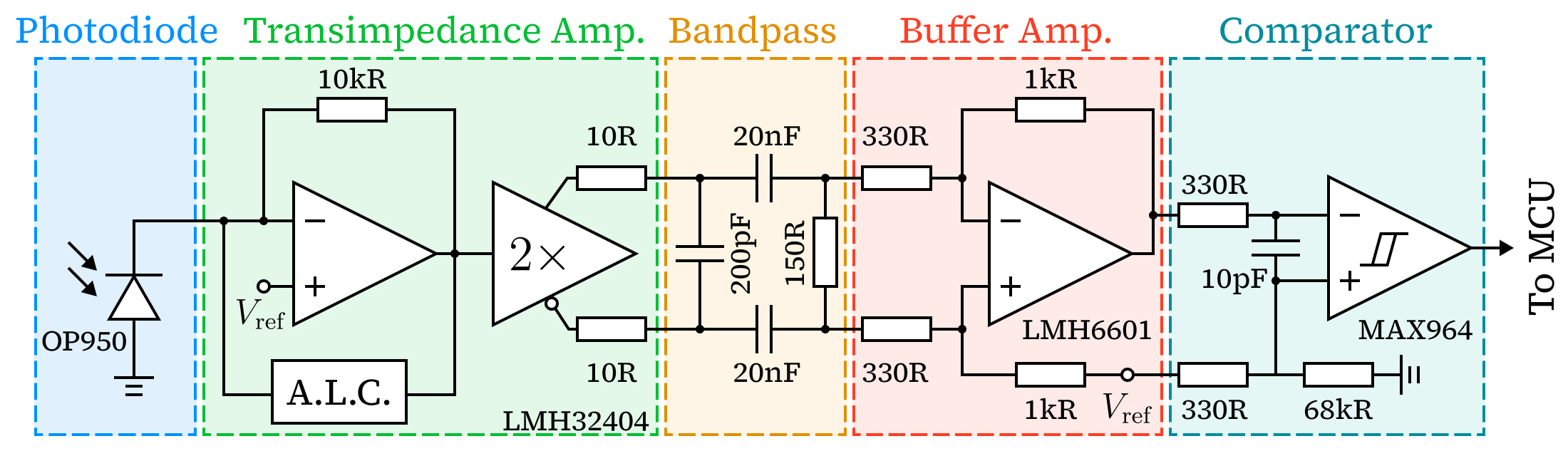}
    \caption{Circuit architecture of the receiver. Amp. and A.L.C. stands for amplifier and ambient light cancellation respectively.}
    \label{fig:circuit}
\end{figure}

The receiver circuit is shown in Fig. \ref{fig:circuit}. PCB layout and filter are designed to reduce electromagnetic interference of the motor and filter out almost all environmental light. The analog part of the circuit has a gain of 67mV/uA, an input referred noise as low as 36nA RMS, and a -3dB passband of [0.42MHz,10.5MHz]. The overall circuit can reliably pickup photocurrent as low as 140nA, \edit{enabling communication across >50cm free space with non-directional transmission and reception.}{enabling all manufactured MP3s to communicate across at least 50cm free space with more than 95\% success rate using non-directional transmission and reception.}

\edit{}{It is worth noting that while the system is capable of communicating over 2m if we remove or enlarge the apertures, we want to limit the communication range to prevent congestion issues in communication. We want drones to be as close to each other as possible to create a dense swarm, so we expect robots to be around 30cm away from each other, and that motivates us to artificially limit the communication range to around $1.5\times$ the average expected distance, or around 50cm.} 

\subsubsection{Communication Protocol Design}
\label{sec:protocol}


When designing the communication protocol, we want it to be general-purpose, to be simple to implement, to have multiplexing ability, and to provide accurate enough timing information $t_{\text{1L}}, t_{\text{1M}},t_{\text{1R}}$ to facilitate position sensing.

To make our protocol general-purpose, we want it to handle messages of different length and content in the same way. This would allow a MP3 to receive communicated positions from other MP3s, as well as other task-dependent messages in future work. A natural approach would be to use packets: a complete message is transmitted in one or multiple 30-bit packets, each containing 2 bytes of data and additional metadata like the ID of the MP3 transmitting the message, the ID of the packet, the cyclic redundancy check (CRC) value of the message, etc. A MP3 can obtain timing information by looking at when it starts to receive packets from another MP3.

We used pulse position modulation (PPM) for transmitting packets because it requires the LED to be on for less amount of time compared to other alternatives like pulse code modulation (PCM) or pulse width modulation (PWM), and PPM can be decoded faster on the microcontroller (ESP32-PICO-D4). Data are modulated in a modified 4-PPM format, where each symbol encodes 2 bits of data and lasts $2^2\times 200$ns, and a 200ns gap is inserted between symbols for the ease of decoding. The top and bottom transmitter start transmission at the same time and transmit the same content, but last pulse duration is 200ns if the packet is sent by the top transmitter and 400ns if sent by the bottom transmitter. This design helps with communication because when a MP3 can see both the top and the bottom transmitter on another MP3, it can still correctly receive the message. Furthermore, during sensing, a MP3 will know whether it is looking at the top or the bottom transmitter on another MP3, and can subsequently compensate for the elevation difference to obtain the actual relative position between MP3s.

Transmitters send packets at an irregular interval \edit{}{to enable pure ALOHA \cite{aloha} multiple access}. \edit{and the}{The} time between two consecutive transmissions is randomly picked \edit{between 50 and 100us}{in the interval of 50\textasciitilde100us} \edit{. This interval is chosen} so that each MP3 is transmitting 20\% of the time on average\edit{ ,and multiple MP3s can communicate with the same receiver using pure ALOHA \cite{aloha} multiple access}{. This means congestion will not be significant when two MP3s are within the communication range and viewing angle of a MP3}. As we have mentioned in section \ref{sec:electrical}, in each revolution a MP3 will have approximately 2\textasciitilde 5ms to communicate with another MP3, and a transmission interval of 75us means an MP3 can transmit about 100 bytes per revolution, \edit{satisfying our requirements}{allowing us to transmit the position estimates of MP3s for localization (6 bytes) while leaving enough capacity for other task-related information. Though limited compared to modern communication systems like WiFi, it is adequate for executing a wide range of swarm algorithms which often do not require high bandwidth between agents}. 

Furthermore, a message interval of 50\textasciitilde 100us means the resolution for timing measurements $t_{\text{1L}}, t_{\text{1M}}, t_{\text{1R}}$ is also 50\textasciitilde 100us, which corresponds to an angle resolution of 0.5\textasciitilde $1^\circ$. Such angle resolution is just enough to obtain useful distance measurements (<2cm RMS error when the distance between MP3s is 30cm). If we increase the FOV of the receivers or reduce the time between transmissions, multiplexing ability will be reduced; and if we decrease the FOV or increase the time between transmissions, MP3s will not be able to transmit as much information in a single rotation. Furthermore, if we increase the time between transmissions, we will also lose position sensing accuracy.

The modified PPM and the parameters we chose for communication also keep the power consumption of transmission in a reasonable range. As the LED is only on for 4\% of the time on average, the total power consumption of the two transmitters is 0.5W, about 5\% of the total power consumption of a flying MP3. 

\subsubsection{Global Localization Algorithm}
\label{sec:algorithm}

We can now bring everything together. In section \ref{sec:theory}, we have already presented the basic theory to use the time when we received messages from other transmitters to determine relative position. The optical and electrical designs (section \ref{sec:optical} and section \ref{sec:electrical}) make MP3s as close to the theoretical model as possible. Finally, we can use the proposed communication system (section \ref{sec:protocol}) to obtain the global position of neighboring MP3s. In this section, we present a simple way of using these relative distance and angle measurements and neighboring MP3s' global position information to determine MP3's global position. We execute localization in three steps to simplify the computation without sacrificing much accuracy:

\begin{enumerate}
    \item Determine the angular velocity $\omega$ using the time when MP3 is facing different transmitters in the last and second last revolution $t_i, t'_i$, where $i=1,2,\dots,N$ are the IDs of the transmitters being seen in both the last and second last revolution.
    \item Determine the global horizontal position $s_X,s_Y$ and the time $t_x$ when MP3 is facing $\hat{x}^+$ direction using angular velocity of the drone $\omega$, the global horizontal position of the transmitters $(\vec{p}_i)_{XY}$, the time when MP3 is facing these transmitters $t_i$, and the measured relative horizontal distance $r_i$.
    \item Determine the global vertical position $s_Z$ using the estimated global horizontal position $s_X,s_Y$, the global position of the transmitters $\vec{p}_i$, and the measured relative elevation angle $\alpha_i$.
\end{enumerate}

In the subsequent discussion, we make the following assumptions:

\begin{enumerate}[label=\alph*)]
    \item MP3's geometric axis is always vertical, and it spins along its geometric axis.
    \item MP3's angular velocity $\omega$ changes slowly (i.e. is constant for a short period of time).
    \item MP3's position changes slowly.
    \item All measurements, including the relative distance measurements $r_i$, relative elevation angle measurements $\alpha_i$, and the time when MP3 is facing other transmitters $t_i$, have Gaussian error distributions and the errors are small.
\end{enumerate}

\noindent The first assumption will introduce error into vertical position estimation as MP3s' geometric axis and angular velocity axis will deviate slightly, but usually they are only off by a few degrees and we are willing to sacrifice the precision in vertical position for faster computation. The second and third assumptions are quite accurate as in experiments the angular velocity changes by <0.5\% and the position usually changes by <5mm in one revolution. While the measured variables are not strictly Gaussian distributed, the fourth assumption is added for ease of explanation without compromising the core idea.

For the first step in the localization procedure, we have

\begin{equation}
    \omega = \frac{2\pi N}{\sum\limits_{i=1}^{N} (t_i-t_i')}
\end{equation}

For the second step, we will determine the most likely $t_x,s_X,s_Y$ given $\vec{p}_i,r_i,t_i$ and their uncertainty estimates $\sigma_{r_i}, \sigma_{t}$.

\begin{equation}
\label{eqn:XYest}
    \mathop{\mathrm{argmin}}\limits_{t_x,s_X,s_Y} \sum\limits_{i=1}^{N} \left(\frac{\Delta_X^2+\Delta_Y^2}{\sigma_{r_i}^2}\right. \left.+\frac{\mathop{\mathrm{atan}}(\Delta_X,\Delta_Y)-\omega(t_i-t_x)}{\omega^2\sigma_{t}^2}\right)
\end{equation}

\noindent where $\Delta_X=(\vec{p}_i)_X-s_X, \Delta_Y=(\vec{p}_i)_Y-s_Y$. We can compute it on the microcontroller using gradient descent at a relative low cost. The covariance matrix $\mathop{\mathrm{Cov}}(s_X, s_Y)$ of $s_X, s_Y$ is given by

\begin{equation}
\label{eqn:uncert1}
    \mathop{\mathrm{Cov}}(s_X, s_Y)=\nabla \nabla f(s_X,s_Y)
\end{equation}

\noindent where $f$ is the summation in (\ref{eqn:XYest}). We can also create a simple measure for the uncertainty of the position estimate as

\begin{equation}
\label{eqn:uncert2}
    \sigma_{XY}=\sqrt{\mathop{\mathrm{Tr}}(\mathop{\mathrm{Cov}}(s_X, s_Y))/2}=\sqrt{\nabla^2 f(s_X,s_Y)/2}
\end{equation}

\noindent which is the root mean square (RMS) of uncertainty in X and Y direction. For simplicity of computation, in the third step we will consider $s_X, s_Y$ as independent random variables that both have uncertainty of $\sigma_{XY}$.

For the third step, we will determine the most likely $s_Z$ using $\vec{p}_i,s_X,s_Y, \alpha_i$ and $\alpha_i$'s uncertainty $\sigma_{\alpha_i}$, but this time a analytical expression could be found:

\begin{equation}
    s_Z=\frac{\sum\limits_{i=0}^{N-1} ((\vec{p}_i)_Z-\sqrt{\Delta_X^2+\Delta_Y^2}\tan \alpha_i)\delta_i^{-2}}{\sum\limits_{i=0}^{N-1} \delta_i^{-2}}
\end{equation}

\noindent where $\delta_i$ is the uncertainty of $(\vec{p}_i)_Z-\sqrt{\Delta_X^2+\Delta_Y^2}\tan \alpha_i$

\begin{equation}
    \delta_i^2 = (\sigma_{XY} \tan\alpha_i)^2+(\sqrt{\Delta_X^2+\Delta_Y^2}\frac{\sigma_{\alpha_i}}{\cos^2 \alpha_i})^2
\end{equation}

\edit{}{Due to manufacturing error of apertures and the placement error of IR photodiodes, while the overall trend given by (\ref{eqn:rh1},\ref{eqn:rh2},\ref{eqn:sh1},\ref{eqn:sh2}) is still valid, the exact value computed using these equations will have significant error. So instead of using these equations directly, we}\edit{We}{} determine the equations for obtaining \edit{$r_i,\alpha_i$ and their uncertainty estimates $\sigma_{r_i},\sigma_{\alpha_i}$}{$r_i,\alpha_i,\sigma_{r_i},\text{ and }\sigma_{\alpha_i}$} from raw timing $t_{\text{iL}},t_{\text{iM}},t_{\text{iR}}$ in a calibration experiment. In the experiment, we place a transmitter at different relative positions to a rotating MP3, record the timing data $t_{\text{iL}},t_{\text{iM}},t_{\text{iR}}$, and find appropriate functions for $r_i,\alpha_i,\sigma_{r_i},\sigma_{\alpha_i}$ that fits the data.

In experiments, we additionally applied an exponential filter with time constant of 0.06s to the obtained position ${s_X,s_Y,s_Z}$ to reduce the noise of position estimates.

\subsection{The Mechanical System}
\label{sec:mechanical}


The overall mechanical design of MP3 is similar to the original Maneuverable Piccolissimo described in \cite{pico}, but with a few changes.

\begin{figure}[htbp]
    \centering
    \includegraphics[width=\linewidth]{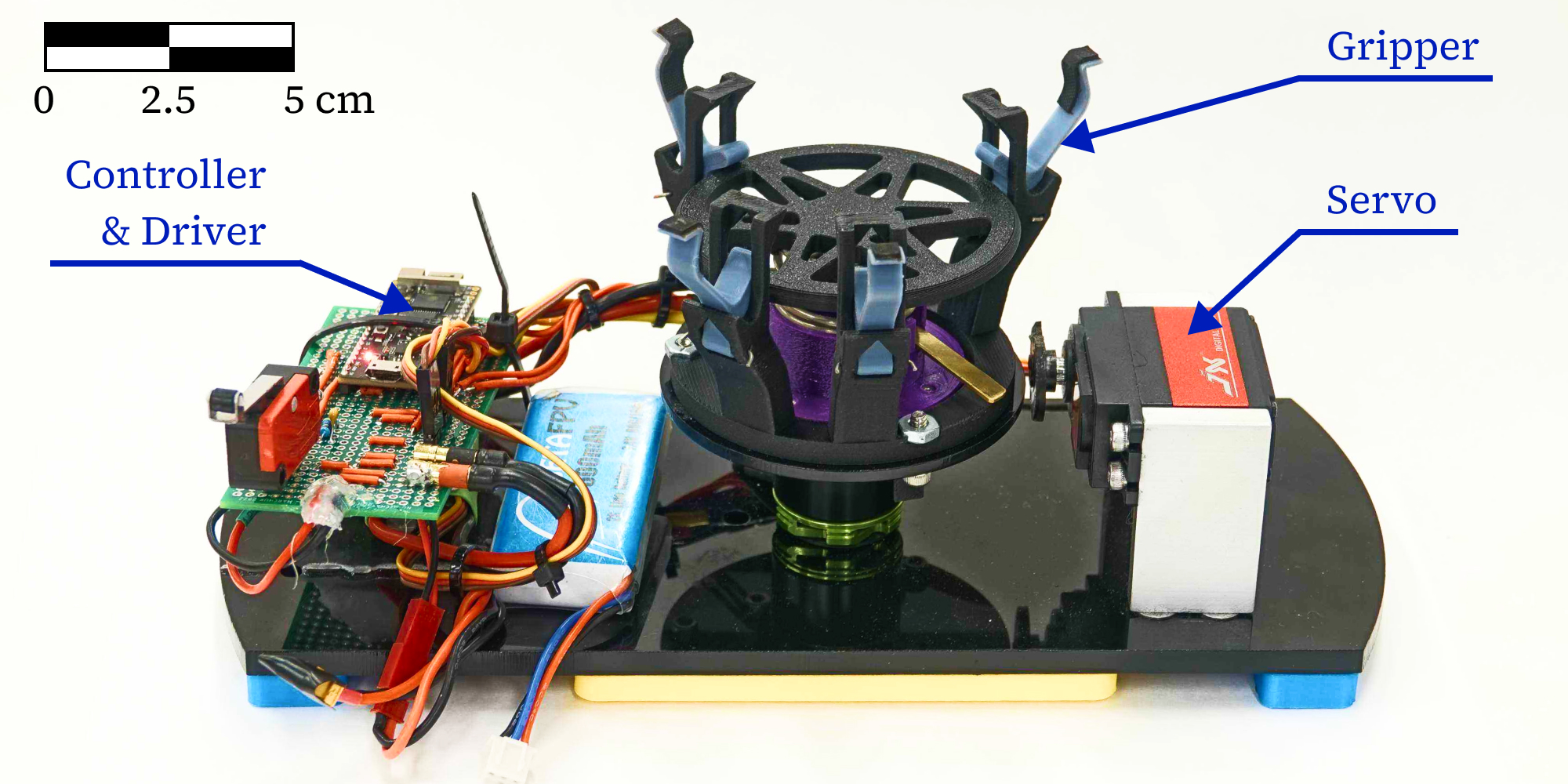}
    \caption{A Photo of the launcher.}
    \label{fig:launcher}
\end{figure}

In MP3, we do not have the mechanism for changing the center of mass during launch like in \cite{drew}, so we used a launcher (shown in Fig. \ref{fig:launcher}) to spin up the drone at takeoff to achieve fast and repeatable launching. The gripper on the launcher seizes the MP3 during the motor's spinning process. Once the MP3 reaches sufficient angular velocity and can maintain stability through the gyroscopic effect, the servo motor on the side \edit{}{of the launcher} will hit a lever to release the key that secures the gripper. Subsequently, the pre-compressed spring will decompress and swiftly disengage the gripper in 5ms, \edit{enabling the MP3 to launch}{launching MP3} into the air.

Several other changes include: a brushless motor instead of a brushed motor to improve longevity and battery life,  carbon fiber-infused nylon for chassis to reduce weight and increase strength, and wings added to the side of MP3 to improve passive stability and reduce rotation rates.

\subsection{The Control Method}
\label{sec:control}

\edit{}{MP3's control method is similar to the method employed in \cite{drew}. Specifically, }MP3 generates two unique motor power commands per \edit{}{drone} rotation to reconcile desired accelerations (both lateral and vertical).\edit{These two unique motor power commands control the tilt of the MP3 which is estimated by the MP3's lateral acceleration.}{ The average of the motor power commands determines the average vertical thrust and vertical acceleration. The difference in motor power commands and the switching time between commands determine the torque acting on MP3 which tilts the drone, resulting in horizontal acceleration. However, the drone does not have independent control of its tilt and position and does not have control over its yaw angle. In addition, because we are controlling MP3's position using only thrust and tilt, the drone is not suitable for very agile and dynamic motions. Despite the limitations in dynamic motion, we believe its control should be sufficient for many swarm behaviors such as studying human-swarm interaction, 3D shape formation, and flocking.} Note that MP3s are spinning too fast for most lightweight IMUs or gyroscopes to operate, so we use lateral acceleration to estimate the tilt.

Despite the overall similarity, a number of details in timing of motor commands, filtering of position, velocity, and acceleration, and the PID parameters have been changed to suit MP3's particular case\edit{ as opposed to the control used in \cite{drew}}{}. Details can be found in the implementation repository \cite{Spinonrepo}.

%
%

\section{Drone Flight Experiments}
\label{sec:droneexp}

We validated the proposed communication and sensing system in several series of experiments. 
\edit{}{In the experiments, we chose the beacons and MP3s' position so they never block the communication of each other and can ensure three beacons or drones are always in sight of each MP3.}

\edit{In the experiments we carried out}{For each experiment}, we collected the reference trajectory of the MP3s using two cameras, one over the apparatus and one on the side, and we also recorded the estimated trajectory of the MP3 which is computed on-board using only relative measurements and data communicated through our IR communication system. Data are recorded after MP3 is flying steadily, about 10\textasciitilde 20s after launch.


\begin{figure}[htbp]
    \centering
    \includegraphics[width=\linewidth]{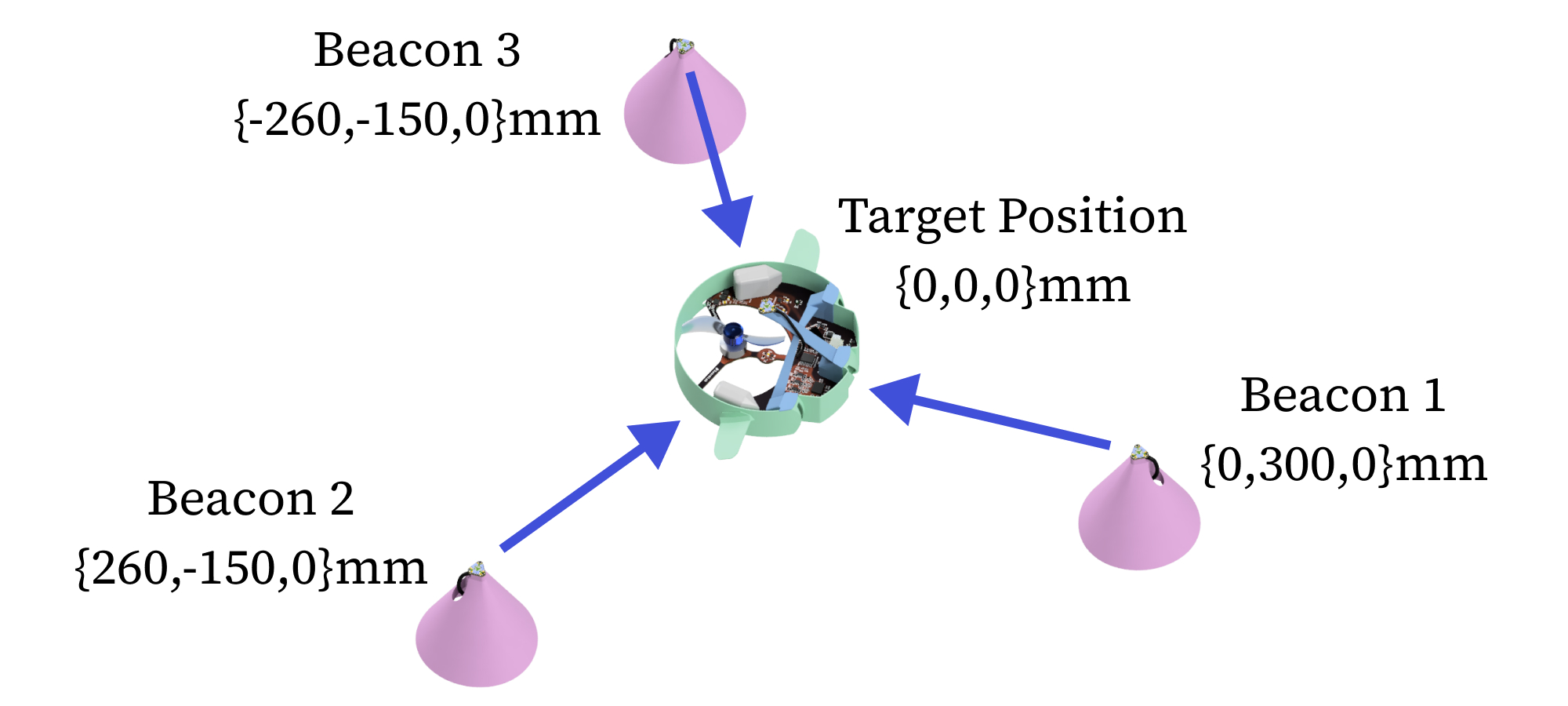}
    \caption{Setup for position holding experiments. Blue arrows mean the beacons can communicate with the MP3. Figure is not drawn to scale.}
    \label{fig:holdsetup}
\end{figure}

\subsection{Position Holding Experiment}
\label{sec:holdexp}

\begin{figure}[htbp]
    \centering
    \includegraphics[width=0.7 \linewidth]{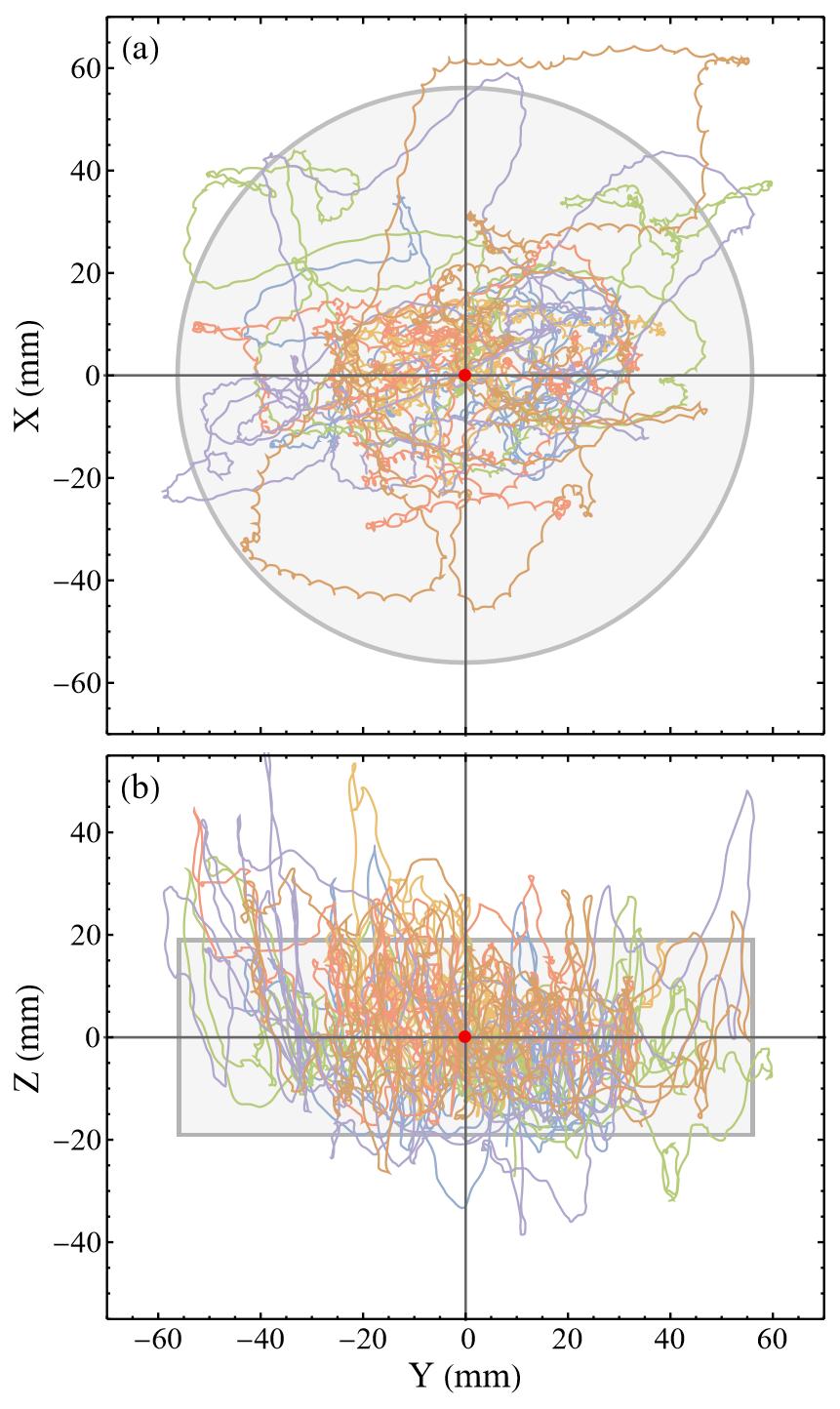}
    \caption{Reference trajectories of MP3 in 6 position holding experiments. Lines of different colors represents trajectories of MP3 in different runs. The gray area's size matches the size of MP3, and the red dots mark the target point.}
    \label{fig:holdtraj}
\end{figure}

In the position holding experiments, MP3 localizes itself by sensing three neighboring beacons. MP3 then tries to remain at a predefined set point at the center of the three beacons. The setup is shown in Fig. \ref{fig:holdsetup}. We plot the reference trajectories of MP3 in all flights in Fig. \ref{fig:holdtraj}. The trajectories indicate successful and stable position holding, since MP3 is staying within 30mm around the origin most of the time (in contrast, the radius of MP3 is 56mm). The root mean square error (RMSE) of the drone's reference position is \{17.6,\,22.5,\,12.7\}mm in X, Y and Z direction respectively.

\begin{figure}[htbp]
    \centering
    \includegraphics[width=0.8\linewidth]{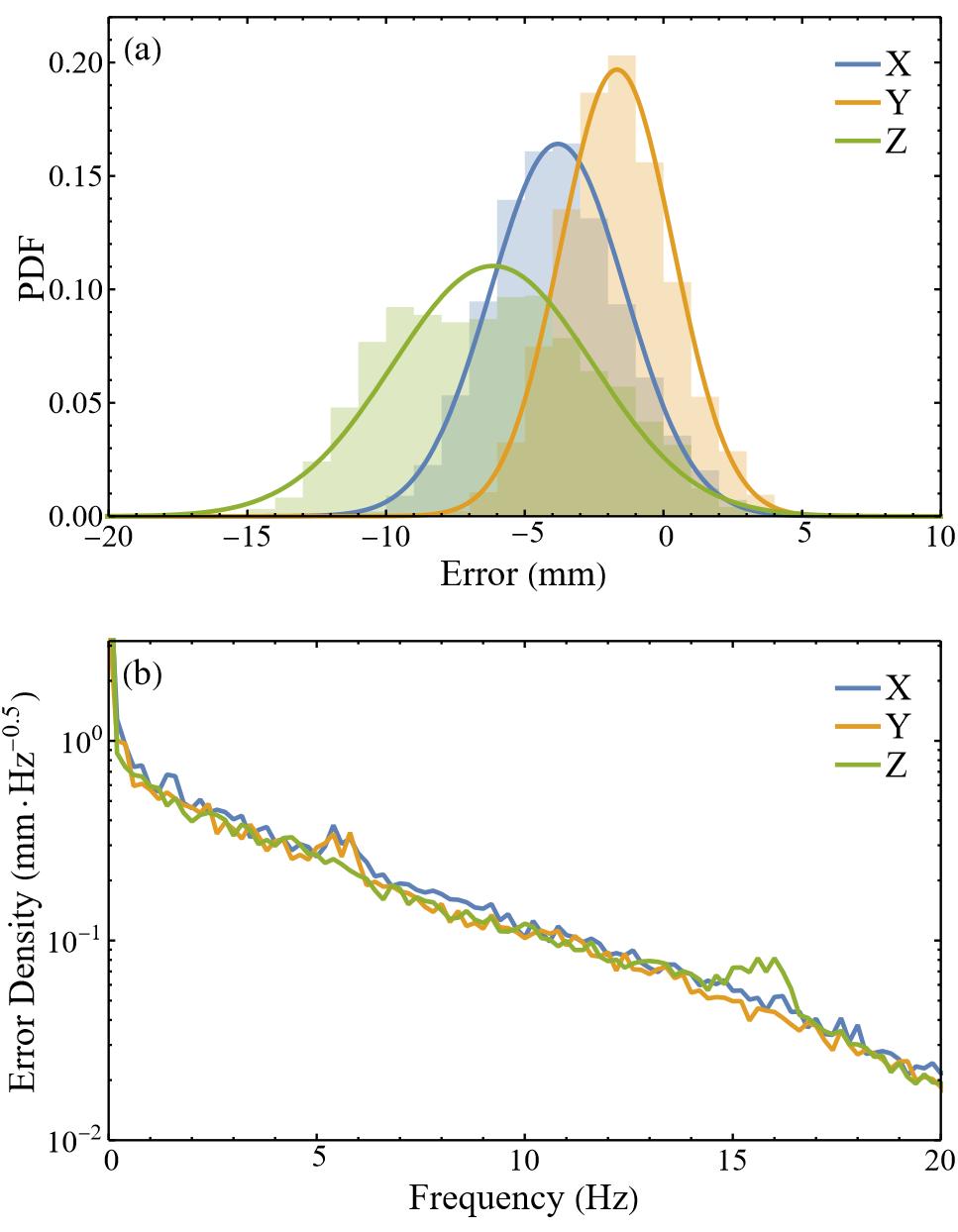}
    \caption{Distribution and spectrum of localization error of MP3 in position holding experiments. (a) shows the localization error's distribution, and (b) shows the localization error's spectrum.}
    \label{fig:holderr}
\end{figure}

A way to evaluate the precision of localization is to calculate the error between the robot-sensed trajectory and reference trajectory. Fig. \ref{fig:holderr}(a) shows the distribution of the localization error in X, Y and Z direction, which resemble normal distribution with $\sigma=\{2.4,2.0,3.6\}$mm respectively. This suggests that our sensing system and localization algorithm can achieve millimeter level accuracy. The distribution's offset in X and Y direction may be due to inaccurate calibration of the camera over the entire flight arena or slight displacement of the beacons, and the offset in Z direction may also be caused by the MP3's tilt.

Besides the localization error's distribution, we are also interested in the spectrum of error, as the higher frequency components in error will be amplified more when we take derivatives to compute the velocity and acceleration. We split the obtained trajectory data into chunks of 5 seconds and resample them at 100Hz. We then compute the discrete Fourier transform (DFT) of the error samples and plot the RMS density of error signal at different frequencies in Fig. \ref{fig:holderr}(b). The lower frequency (<2Hz) components are dominated by systematic error like the inaccurate placement of beacons or inaccuracy in the functions that converts time $t_{1L},t_{1M},t_{1R}$ to distance and elevation angle. These sources are dependent on the position of the drone which is slowly changing. The higher frequency (>2Hz) components, however, mainly come from the discrete and random nature of timing measurements $t_{1L},t_{1M},t_{1R}$. Because the error of timing measurement is different and nearly independent in every relative distance measurement, the spectrum of the position error caused by this factor will spread to higher frequency range.

The high-frequency error component significantly effects the drone's stability during hovering. When it is much higher, it will introduce a large error in velocity and acceleration estimates and saturate the PID controller, causing the drone to lose control.

\begin{figure}[tbph]
    \centering
    \includegraphics[width=\linewidth]{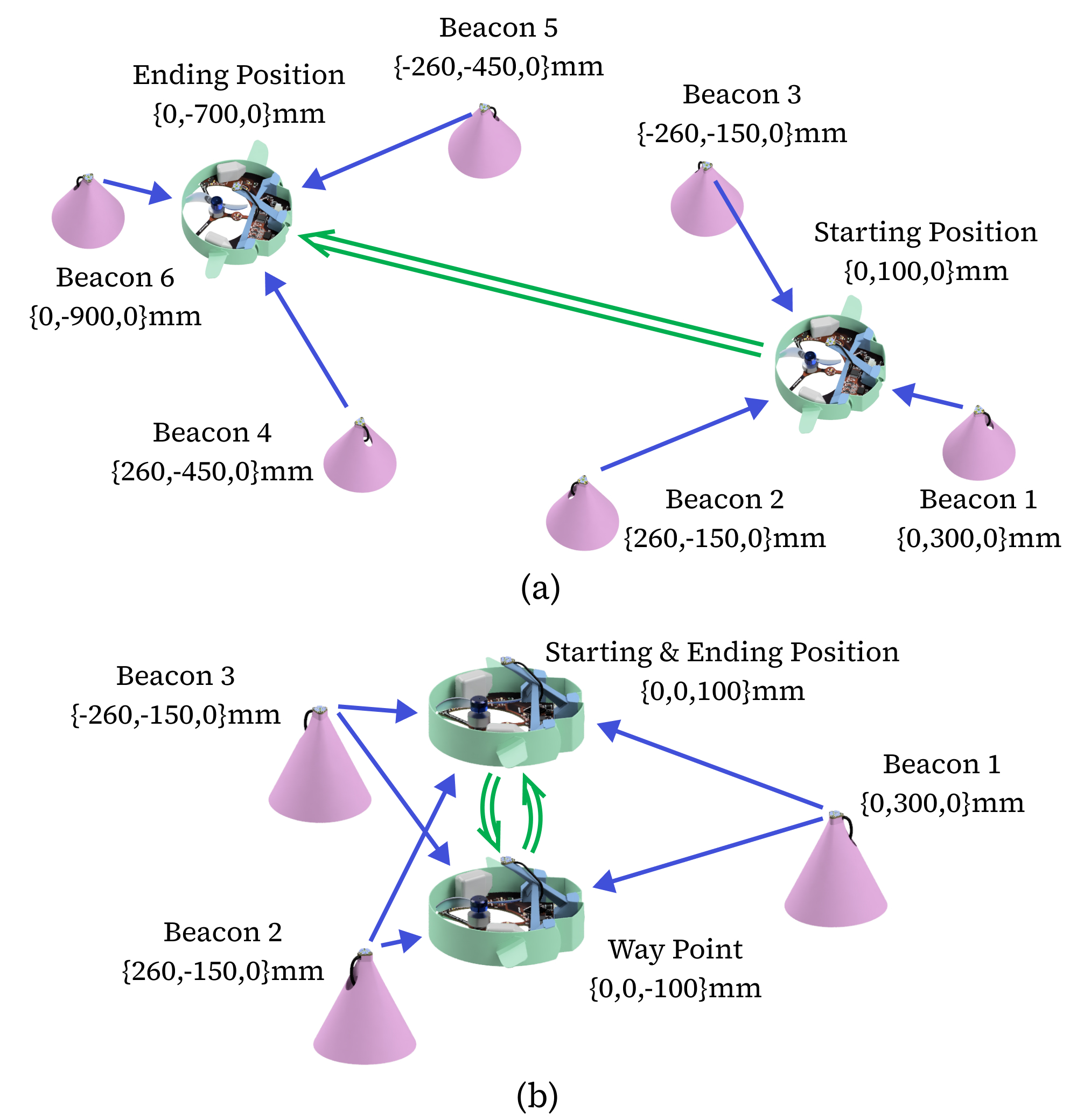}
    \caption{Setup for horizontal and vertical movement experiments. (a) shows the setup for the horizontal movement experiment, and (b) shows the setup for the vertical movement experiment. Blue arrows mean the beacons can communicate with the MP3, and green arrows show the target movement trajectory. Figure is not drawn to scale.}
    \label{fig:movesetup}
\end{figure}

\subsection{Horizontal and Vertical Movement Experiment}
\label{sec:stepexp}

\begin{figure}[htbp]
    \centering
    \includegraphics[width=0.8 \linewidth]{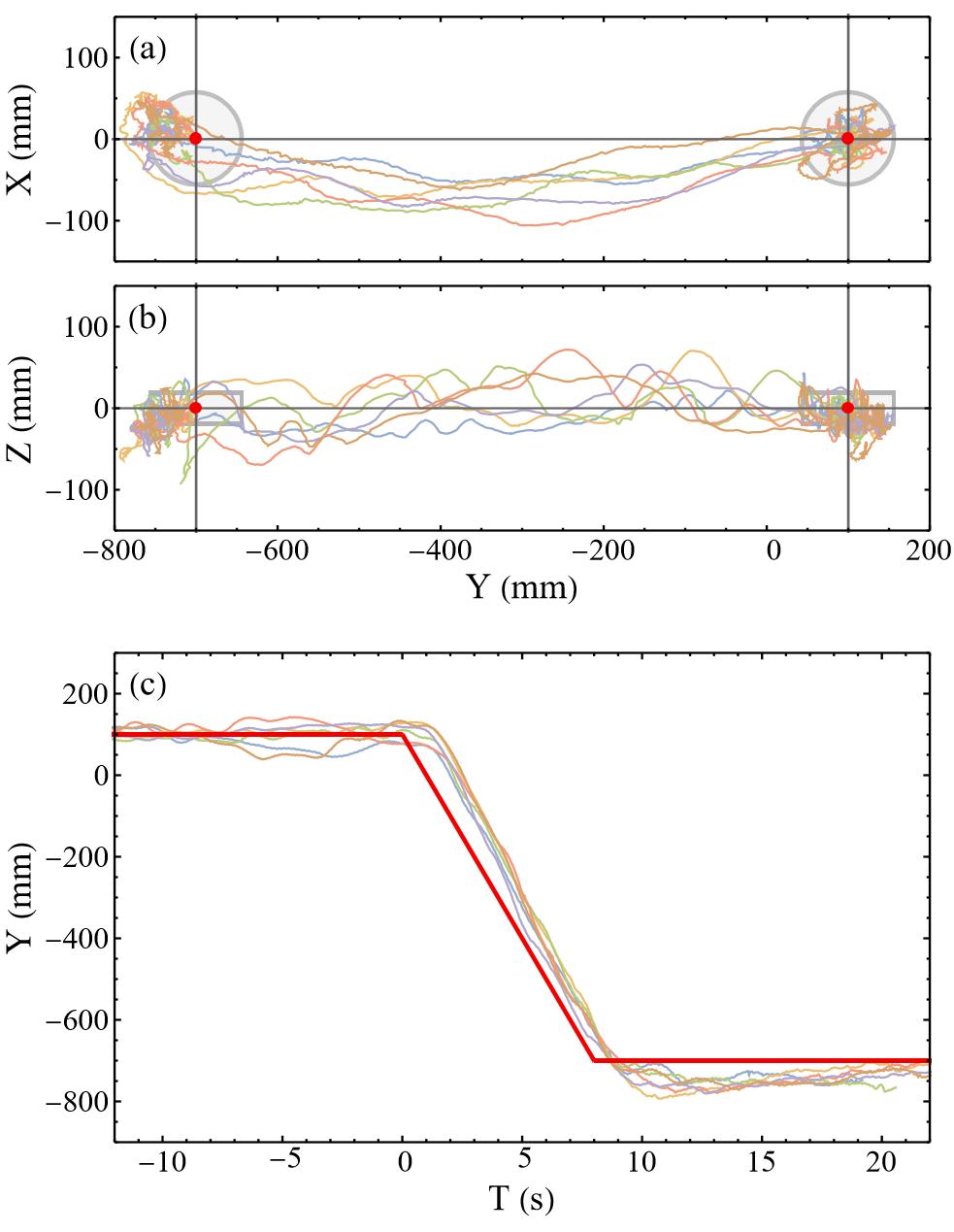}
    \caption{Reference trajectories of MP3 in 6 horizontal movement experiments. (a) and (b) show the trajectory, and (c) shows the Y-T plot where the thick red line is the target trajectory. }
    \label{fig:hortraj}
\end{figure}

\begin{figure}[htbp]
    \centering
    \includegraphics[width=0.8 \linewidth]{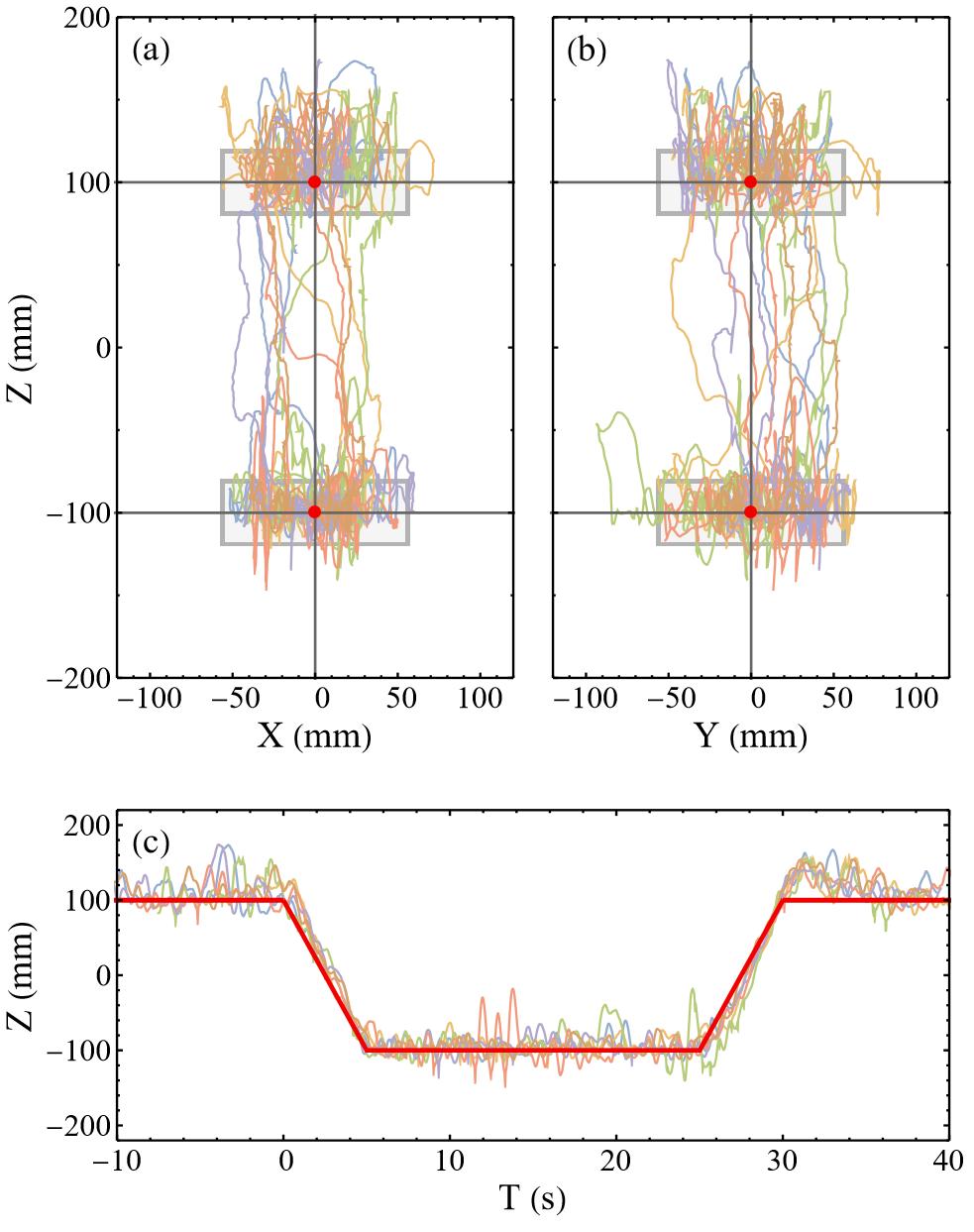}
    \caption{Reference trajectories of MP3 in 6 vertical movement experiments. (a) and (b) show the trajectory, and (c) shows the Z-T plot where the thick red line is the target trajectory.}
    \label{fig:vertraj}
\end{figure}

In the movement experiments, instead of remaining at a position, MP3 moves between way points. The setup is shown in Fig. \ref{fig:movesetup}.
MP3's reference trajectory and its Y coordinate versus time in the 6 horizontal movement experiments are shown in Fig. \ref{fig:hortraj}. MP3's reference trajectory and its Z coordinate versus time in the 6 vertical movement experiments are shown in Fig. \ref{fig:vertraj}. The experiments clearly show that MP3 is capable of controlled motion in 3D using the on-board communication and sensing system.

\begin{figure}[htbp]
    \centering
    \includegraphics[width=\linewidth]{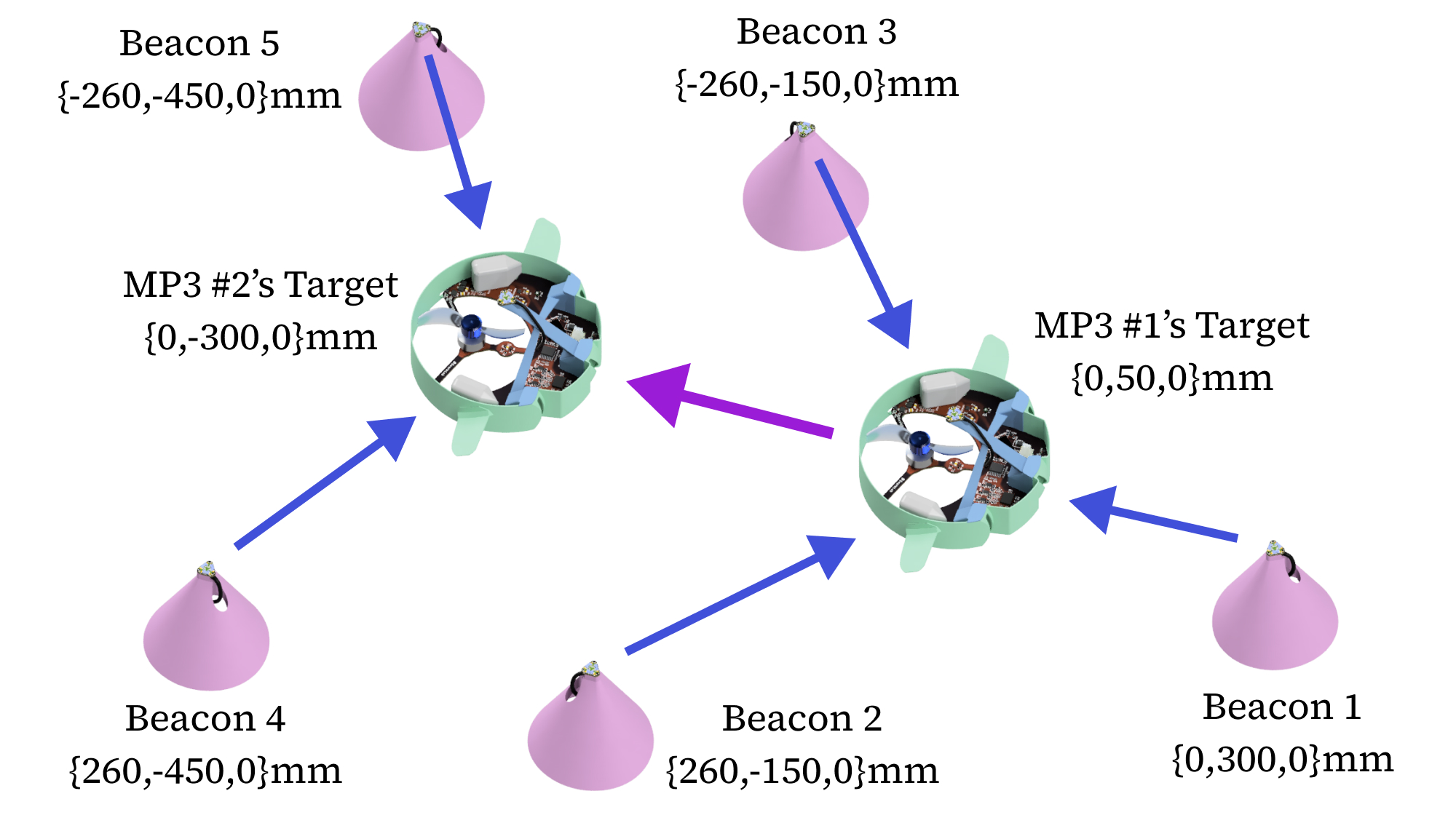}
    \caption{Setup for peer-to-peer communication and sensing experiments. Blue arrows mean the beacons can communicate with the MP3, and purple arrows means the MP3 can communicate with its peers. Figure is not drawn to scale.}
    \label{fig:P2Psetup}
\end{figure}

\subsection{Peer-to-Peer Communication and Sensing Experiment}
\label{sec:p2pexp}

\begin{figure}[!t]
    \centering
    \includegraphics[width=0.9 \linewidth]{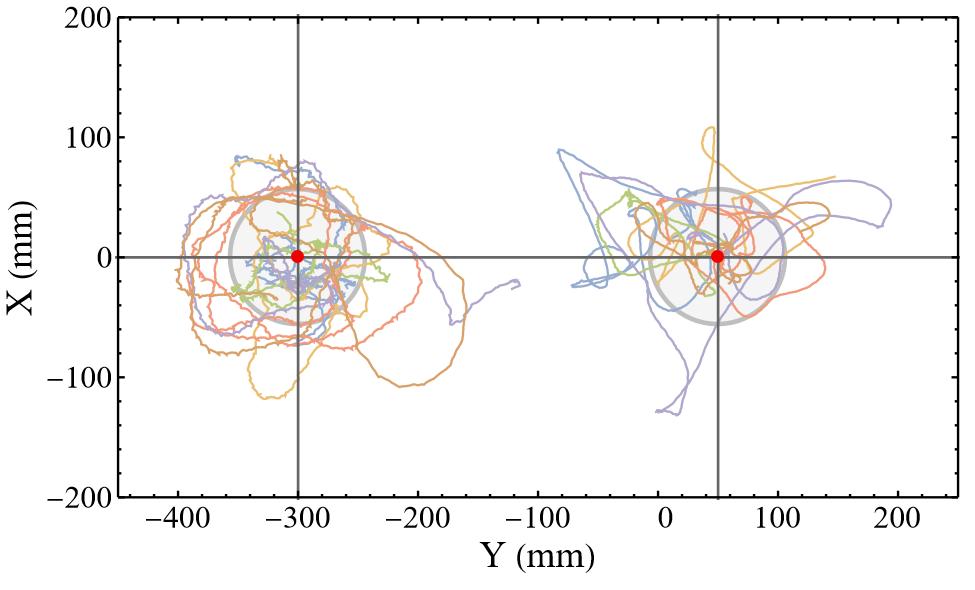}
    \caption{Reference trajectories of MP3 \#1 and MP3 \#2 in 6 peer-to-peer communication and sensing experiments. The MP3 on the right and left are MP3 \#1 and MP3 \#2 respectively, and their respective trajectories in a single run are shown in the same color.}
    \label{fig:P2Ptraj}
\end{figure}

In this set of experiments, we demonstrate the capability of MP3 to do peer-to-peer communication and sensing. The setup is shown in Fig. \ref{fig:P2Psetup}. Unlike in previous experiments, MP3 \#1 is actively transmitting its current sensed position to MP3 \#2, acting like a beacon. In software, we made sure that MP3 \#1 can only communicate with the first three beacons, and MP3 \#2 can only communicate with the last two beacons and MP3 \#1. MP3 \#2 will record two sets of position estimates, one using both the information from the beacons and the relative measurements and position estimates of MP3 \#1, and another using only the information from the beacons. The horizontal reference trajectory of MP3s in all 7 flights are shown in Fig. \ref{fig:P2Ptraj}. 

\begin{figure}[!t]
    \centering
    \includegraphics[width=0.9 \linewidth]{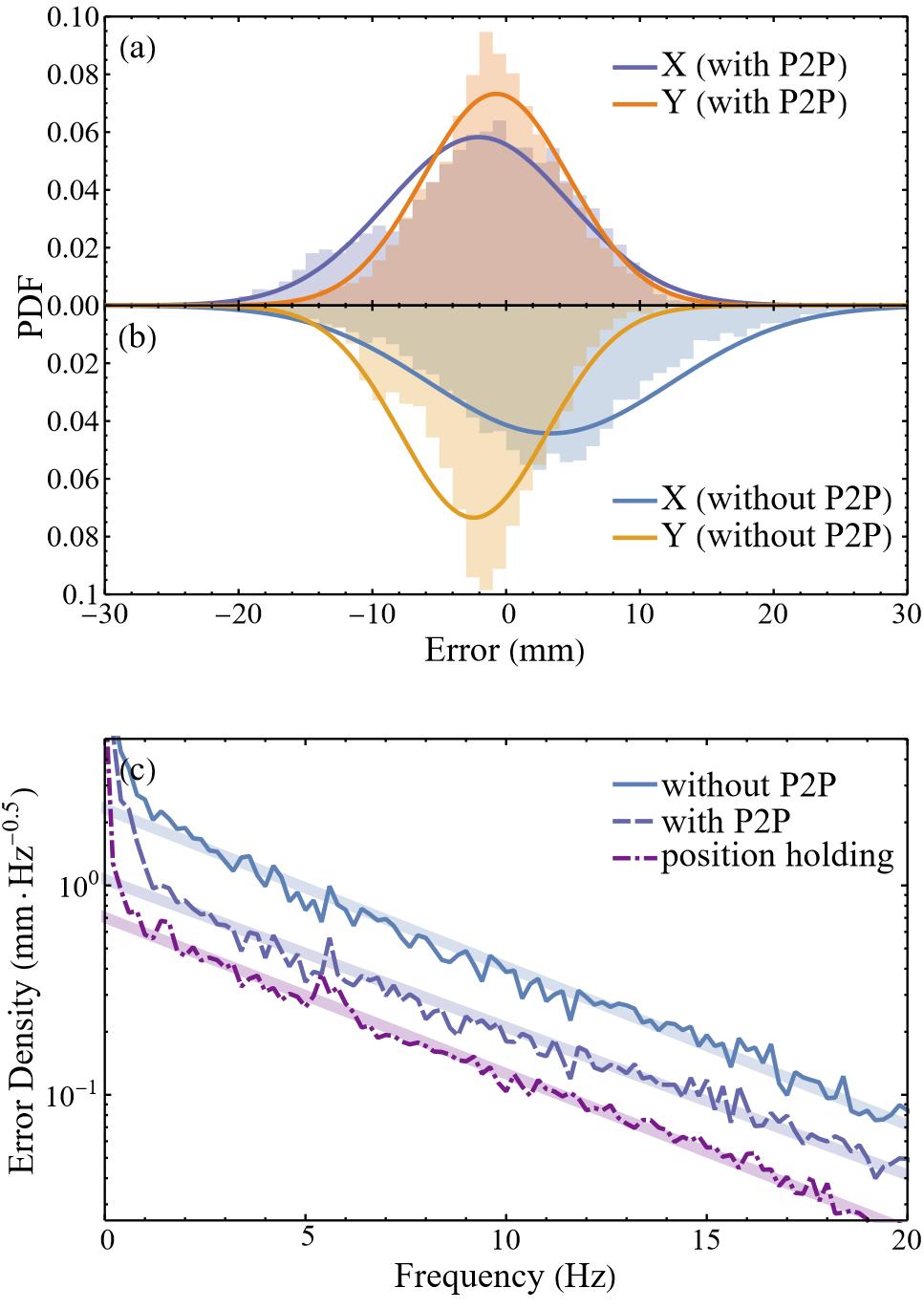}
    \caption{Distribution and spectrum of localization error of MP3 \#2 in peer-to-peer communication and localization experiments. (a) and (b) show the localization error's distribution where MP3 \#2 has used and not used peer-to-peer information for localization respectively. (c) compares the Y direction localization error's spectrum in different scenarios. The purple dot dashed line in (c) shows the spectrum of Y localization error in position holding experiment (Fig. \ref{fig:holderr}(b)), and the thicker lines in the background are the linear fits of high frequency components. }
    \label{fig:P2Perr}
\end{figure}

We are especially interested in the flight performance of MP3 \#2 horizontally, which most directly shows the effect of the information provided by MP3 \#1. The localization error's distribution in X and Y direction with or without peer-to-peer information is shown in Fig. \ref{fig:P2Perr}(a) and (b). While the peer-to-peer communication has little impact on the localization accuracy in Y direction, it reduces the RMS error of localization in X direction by approximately 50\%. This result is consistent with the uncertainty given by (\ref{eqn:uncert1}).

We could also plot the spectrum of localization error in X direction in Fig. \ref{fig:P2Perr}(c). It shows that the peer-to-peer communication and sensing with MP3 \#1 helped to reduce the high frequency component by 6dB or 50\%. This reduction in high frequency error is crucial for MP3 \#2 to stay stable.\\

In fact, if there is no MP3 \#1 and MP3 \#2 is only localizing itself using two beacons, MP3 \#2 will soon lose position stability due to the error in velocity and acceleration estimations. 


\section{Discussion}
\label{sec:discussion}


The MP3 prototype is designed to test the feasibility of the communication, sensing, localization, and flight control system. There are a few aspects that could improve the overall performance of MP3 and enable MP3 to form larger swarms.

\edit{}{One key improvement could be the flight time. Currently, the flight time of MP3 is approximately 3min and is limited by the operating voltage of the motor driver. However, the design of MP3 was not optimized for the flight time, and we anticipate that the flight time could be extended to 5-6min given small changes in motor driver and power circuit.}

Another key improvement could be the communication system. Currently, we are using pulse position modulation and ALOHA multiple access for communication, which has low bandwidth, is less tolerant to error, and has congestion issues when more than two transmitters are trying to transmit information to the same receiver. These issues could be addressed using techniques like quadrature amplitude modulation (QAM) and frequency division multiple access (FDMA), though at a cost of hardware complexity. A simpler approach might be slotted ALOHA, which could relieve the congestion issue and approximately double the total bandwidth, but would require a time synchronization between MP3s.

Improvement could also be made to the localization and control system. \edit{Having }{We can create a more accurate sensing system by reducing $\omega\sigma_t$. This can be accomplished by reducing the rotation speed of the drone or increasing the frequency of communication. In addition, we can improve the state estimation and control by having }a better dynamic model of MP3 and higher resolution motor speed feedback.\edit{ could lead to better state estimation and control}{}.

We have also made some preliminary progress in enabling MP3 to actively sense the environment. Light transmitted by MP3 will be reflected by the objects in the environment, and then captured by the receivers on the same MP3. Objects reflecting the light are very similar to transmitters, and we can sense their relative position to the MP3 in the way similar to how we sense the relative position of other transmitters. In certain environments, it may be possible to achieve solo MP3 flights with just reflected light information.

\section{Conclusion} 
\label{sec:conclusion}

In this paper, we presented MP3, a minimalist, single propeller drone equipped with novel peer-to-peer communication and sensing mechanisms. We explained the hardware and software architecture of the drone, especially its communication and sensing system, and demonstrated its capability to localize itself by communicating with and sensing its relative bearing, distance, and elevation to neighboring MP3s. We also showed that MP3 is capable of flying stably alone as well as together with peers. Thanks to the single-motor design and the novel communication and sensing system, MP3 is about 50\% lighter than the lightest drone known to the authors that has relative position sensing capabilities of similar accuracy \cite{light1}. Having these three key capabilities, MP3 is theoretically capable to execute a wide variety of swarm tasks like collaborative search or shape formation in a fully distributed manner. 

\section*{Acknowledgments}

Thanks to Marko Vejnovic for reviewing the program's architecture and providing helpful feedback.  This work was supported by The National Science Foundation, NRI2.0 grants 2024692 and 2024615.

\bibliographystyle{unsrtnat}
\bibliography{references}

\end{document}